\documentclass[10pt, twocolumn, journal]{IEEEtran}

\renewcommand{\baselinestretch}{1.0} % Change to 1.65 for double spacing

\usepackage[pdftex]{graphicx,xcolor}

\usepackage{amsmath}
\usepackage{cite}
\usepackage{booktabs}
\usepackage{multirow}
\usepackage{tabularx}
\usepackage{url}
\usepackage{adjustbox}
\usepackage{array}
\usepackage{overpic}
\usepackage{colortbl}
\usepackage[colorlinks=true, allcolors=blue]{hyperref}
\usepackage{ulem}
\usepackage{setspace}
\usepackage{float} % stop figure from floating with [H] (package {here} is obsolete
\usepackage{subfig}
\usepackage{hyperref}
\usepackage{soul}
\usepackage{wrapfig} % for author biographies

\newcommand{\ifmiaAuthorAff}[2]{${}^{#1}$\normalsize{\it{#2}}}

\newcommand{\ifmiaKeywords}[1]{\vspace{1mm} \noindent \small{{\bf Keywords: } #1}}

\renewcommand{\thetable}{\arabic{table}}

\begin{document}
%%%%%%%%%%%%%%%%%%%%%%%%%%%%%%%%%%%%%%%%%%%%%%%%%%%%%%%%%%%%%%%%%%%%%%%%%%%%%%%%%%%%%%%%%%%%%%%%
%\title{\Large{Dense volumetric medical image segmentation using 3D fully convolutional networks}}
\title{\Large{Deep learning and its application to medical image segmentation}}
%%%%%%%%%%%%%%%%%%%%%%%%%%%%%%%%%%%%%%%%%%%%%%%%%%%%%%%%%%%%%%%%%%%%%%%%%%%%%%%%%%%%%%%%%%%%%%%%
%%%author list
\author{Holger R. ROTH\ifmiaAuthorAff{1},
Chen SHEN\ifmiaAuthorAff{1},
Hirohisa ODA\ifmiaAuthorAff{2},
Masahiro ODA\ifmiaAuthorAff{1},
Yuichiro HAYASHI\ifmiaAuthorAff{1},
Kazunari MISAWA\ifmiaAuthorAff{3},\\
Kensaku MORI\ifmiaAuthorAff{1}\\
\vspace{3mm}
	
\ifmiaAuthorAff{1}{Graduate School of Informatics, Nagoya University, Japan}\\
\ifmiaAuthorAff{2}{Graduate School of Information Science, Nagoya University, Japan}\\
\ifmiaAuthorAff{3}{Aichi Cancer Center Hospital, Japan}

\vspace{-7mm}}% <-this % stops a space


\maketitle


\begin{abstract}
One of the most common tasks in medical imaging is semantic segmentation. Achieving this segmentation automatically has been an active area of research, but the task has been proven very challenging due to the large variation of anatomy across different patients. However, recent advances in deep learning have made it possible to significantly improve the performance of image recognition and semantic segmentation methods in the field of computer vision. Due to the data driven approaches of hierarchical feature learning in deep learning frameworks, these advances can be translated to medical images without much difficulty. Several variations of deep convolutional neural networks have been successfully applied to medical images. Especially fully convolutional architectures have been proven efficient for segmentation of 3D medical images. In this article, we describe how to build a 3D fully convolutional network (FCN) that can process 3D images in order to produce automatic semantic segmentations. The model is trained and evaluated on a clinical computed tomography (CT) dataset and shows state-of-the-art performance in multi-organ segmentation.
\end{abstract}

\ifmiaKeywords{fully convolutional networks, deep learning, segmentation, computed tomography}

%%%%%%%%%%%%%%%%%%%%%%%%%%%%%%%%%%%%%%%%%%%%%%%%%%%%%%%%%%%%%%%%%%%%%%%%%%%%%%%%%%%%%%%%%%%%%%%%
%%%%%%%%%%%%%%%%%%%%%%%%%%%%%%%%%%%%%%%%%%%%%%%%%%%%%%%%%%%%%%%%%%%%%%%%%%%%%%%%%%%%%%%%%%%%%%%%
\section{INTRODUCTION}
Automated segmentation of medical images is challenging because of the large shape and size variations of anatomy between patients. Furthermore, low contrast to surrounding tissues can make automated segmentation difficult \cite{roth2017spatial}. Recent advantages in this field have mainly been due to the application of deep learning based methods that allow the efficient learning of features directly from the imaging data. Especially, the development of fully convolutional neural networks (FCN) \cite{long2015fully} has further improved the state-of-the-art in semantic segmentation of medical images.

In this article, we describe how FCNs have been derived from CNNs and how to utilize 3D FCNs that can segment volumetric medical images with high accuracy and robustness.

\subsection{Convolutional neural networks (CNN)}
Many of the recent advances in computer visions are due to the efficient application of convolutional neural networks (CNN) on  graphics processing units (GPUs). GPU acceleration has significantly sped up computations, allowing the training of very deep and complex models on large datasets. For example, Krizhevsky et al. \cite{krizhevsky2012imagenet} nearly halved the error rate on the \textit{ImageNet} challenge dataset from one year to the next by training a deep CNN on two GPU cards on over one million images. CNNs are effective because they can learn hierarchical feature representations of the image in a purely data-driven manner. This means that features which are good for classification are learned from the images just given a supervisory signal that defines the desired classification output. This so-called ``supervised learning'' has been recently applied to many fields of science, including biomedical and radiological imaging \cite{cruz2013deep,roth2014new,shin2016deep,kamnitsas2017efficient}, and significantly advanced the state-of-the-art \cite{greenspan2016guest}. 

Typically, a CNN consists of several layers of convolutional, pooling, and fully-connected (or densely connected) neural network layers \cite{lecun2015deep}. The convolutional layers make use of spatial correlation in the input images by sharing the filter kernel weights for the computation of each feature map. Pooling layers allow reducing the dimensions of each input feature map while preserving the most relevant feature responses. Commonly used pooling includes max- or average-pooling. Max-pooling can also add some invariance to local shifting of the objects in the input image. The outputs of each CNN layer are typically fed to non-linear activation functions (often rectified linear units (ReLUs) \cite{krizhevsky2012imagenet}). The use of non-linear activation functions allows us to model very complex mappings between the input image and the desired outputs.

Figure \ref{fig:cnn} shows a schematic overview of a typical CNN architecture that produces a per-image prediction by using a \textit{softmax} output for multi-class classification. This type of architecture has been successfully applied to many medical imaging tasks. A few examples in radiology are anatomy classification \cite{roth2015anatomy,yan2016multi}, false-positives reduction for computer-aided detection of lymph nodes or colonic polyps \cite{roth2016improving}, the detection of pulmonary nodules \cite{setio2016pulmonary}, and the detection of pulmonary embolisms \cite{tajbakhsh2015computer}. CNNs have also been successfully applied to the classification of endoscopic video sequences, e.g. in colonoscopy \cite{tajbakhsh2016convolutional} or laparoscopic surgery \cite{twinanda2017endonet}.

\begin{figure*}[htb]
	\centering
	\adjincludegraphics[width=0.8\textwidth]{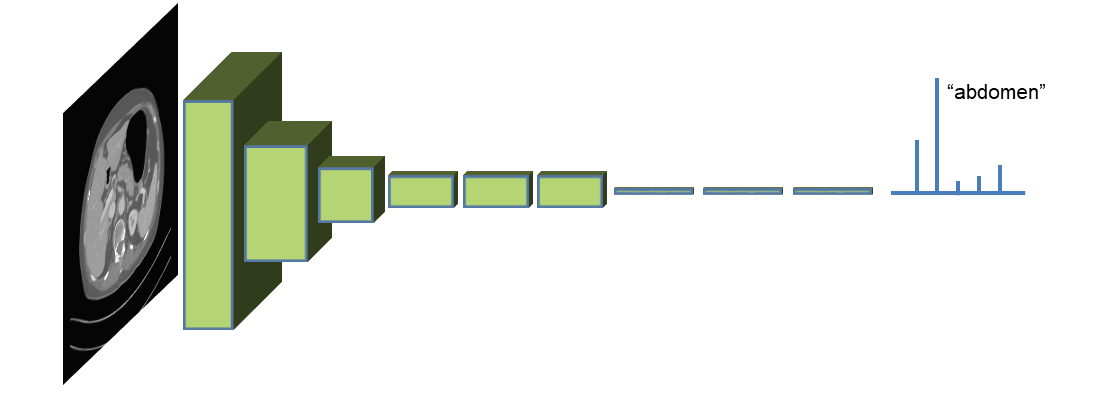}
	\caption{Convolutional neural network (CNN). A straightforward application of CNNs for anatomy classification in whole body CT scans can be found in \cite{roth2015anatomy} (illustration after \cite{long2015fully}).}
	\label{fig:cnn}
\end{figure*}
%Use in medical imaging. 2.5D approach, 3D CNN (deep medic, hong kong?)

\subsection{Fully convolutional networks (FCN)}
One downside of CNNs is that the spatial information of the image is lost when the convolutional features are fed into the final fully connected layers of the network. However, spatial information is especially important for semantic segmentation tasks. Hence, the fully convolutional network (FCN) was proposed by Long et al. \cite{long2015fully} to overcome this limitation. In FCNs, the final densely connected layers of the CNN are replaced by transposed convolutional layers in order to apply a learned up-sampling to the low-resolution feature maps within the network. This operation can recover the original spatial dimensions of the input image while performing semantic segmentation at the same time. Similar network structures have been successfully applied to semantic segmentation tasks in medical imaging \cite{roth2016spatial,zhou2016three,zhou2017fixed} and to the segmentation of biomedical images such as histology slides \cite{ronneberger2015unet}. Extensions to 3D biomedical imaging data from modalities such as confocal microscopy \cite{cciccek20163d} or magnetic resonance imaging (MRI) have been proposed \cite{milletari2016v}. In a typical FCN architecture, skip connections can be utilized to connect different levels of the network in order to preserve image features that are ``closer'' to the original image. This helps the network to achieve a more detailed segmentation result and can simplify or speed up training \cite{drozdzal2016importance,szegedy2017inception}. The typical setup of an FCN is illustrated in Fig. \ref{fig:fcn} for the semantic segmentation of CT image slices.
%%%%%%%%%%%%%%%%%%%%%%%%%%%%%%%%%%%%%%%%%%%%%%%%%%%%%%%%%%%%%%%%%%%%%%%%%%%%%%%%%%%%%%%%%%%%%%%%
\begin{figure*}[htb]
	\centering
	\adjincludegraphics[width=0.8\textwidth]{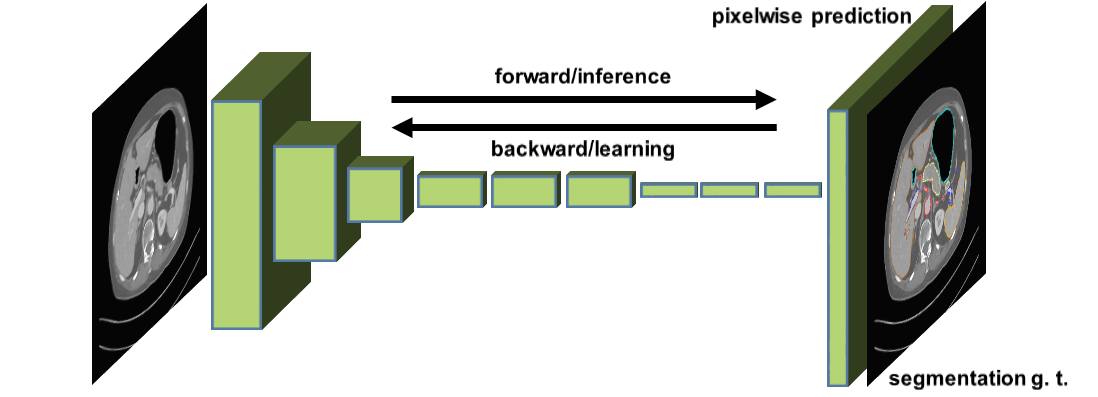}
	\caption[FCN]{Fully convolutional network (FCN). Examples of FCNs applied to semantic segmentation tasks in medical imaging can be found in \cite{roth2016spatial,zhou2016three,zhou2017fixed,zhou2017deep} (illustration after \cite{long2015fully}).}
	\label{fig:fcn}	
\end{figure*}
%%%%%%%%%%%%%%%%%%%%%%%%%%%%%%%%%%%%%%%%%%%%%%%%%%%%%%%%%%%%%%%%%%%%%%%%%%%%%%%%%%%%%%%%%%%%%%%%

\subsection{Related work}
Semantic segmentation of CT images has been an active area of research over the years. Classical approaches for multi-organ segmentation range from statistical shape models \cite{cerrolaza2015automatic,okada2015abdominal} to techniques that employ image registration. Many methods include some form of multi-atlas label fusion \cite{rohlfing2004evaluation,wang2013multi,iglesias2015multi} which has been widely applied in clinical research and practice. Furthermore, approaches that combine techniques from multi-atlas registration and machine learning have been proposed \cite{tong2015discriminative,oda2016regression}. However, the difficulties of modeling the complex shape variations between patients, especially for abdominal organs, have made it difficult for registration-based methods to perform adequately for very non-rigid organs \cite{lee2015evaluation}. 
	
Today, many successful deep learning methods from computer vision are being adapted to segmentation tasks in medical imaging. Recent examples include \cite{roth2015deeporgan,roth2017spatial,zhou2016pancreas,christ2016automatic,zhou2016three,milletari2016v,christ2016automatic,kamnitsas2017efficient} and many others. Most of these methods are based on 2D and 3D variants of FCNs \cite{long2015fully} that allow the extraction of features that are useful for image segmentation directly from the imaging data. This is crucial for the success of deep learning \cite{lecun2015deep} and avoids the need for ``hand-crafting'' features suitable for detection of individual organs.
%%%%%%%%%%%%%%%%%%%%%%%%%%%%%%%%%%%%%%%%%%%%%%%%%%%%%%%%%%%%%%%%%%%%%%%%%%%%%%%%%%%%%%%%%%%%%%%%
%%%%%%%%%%%%%%%%%%%%%%%%%%%%%%%%%%%%%%%%%%%%%%%%%%%%%%%%%%%%%%%%%%%%%%%%%%%%%%%%%%%%%%%%%%%%%%%%
\section{METHODS}
%%%%%%%%%%%%%%%%%%%%%%%%%%%%%%%%%%%%%%%%%%%%%%%%%%%%%%%%%%%%%%%%%%%%%%%%%%%%%%%%%%%%%%%%%%%%%%%%
\begin{figure*}[htb]
	\centering
	\adjincludegraphics[width=0.95\textwidth]{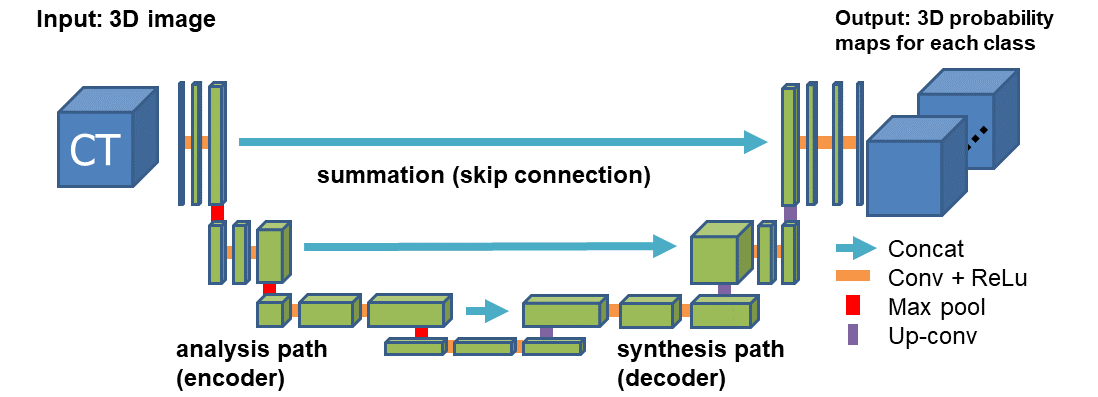}
	\caption{The architecture of our 3D U-Net like fully convolutional network. It applies an end-to-end architecture using same size convolutions (via zero padding) with kernel sizes of $3\times3\times3$. \label{fig:3Dunet}}
\end{figure*}
%%%%%%%%%%%%%%%%%%%%%%%%%%%%%%%%%%%%%%%%%%%%%%%%%%%%%%%%%%%%%%%%%%%%%%%%%%%%%%%%%%%%%%%%%%%%%%%%
FCNs have made it feasible to train models for pixel-wise semantic segmentation in an end-to-end fashion \cite{long2015fully}. Efficient implementations of 3D convolution and growing GPU memory have further made it possible to extend these methods to 3D medical imaging and to train networks on large amounts of annotated image volumes. One such example is the recently proposed 3D U-Net architecture \cite{cciccek20163d}. In the following, we describe how to apply the 3D U-Net architecture to the problem of multi-organ segmentation in CT. 
%%%%%%%%%%%%%%%%%%%%%%%%%%%%%%%%%%%%%%%%%%%%%%%%%%%%%%%%%%%%%%%%%%%
\subsection{3D Fully convolutional networks (3D U-Net)} As described above, FCNs have the ability to solve challenging classification tasks in a data-driven manner, given a training set of images and labels $S={(I_n,L_n )}$ with $n=1,\dots,N$, where $I_n$ denotes one of $N$ CT images and $L_n$ denotes the corresponding ground truth label image. This setup allows the network to find a direct mapping from the image to the segmentation by learning a very complex non-linear function between the two. The 3D U-Net architecture \cite{cciccek20163d} consists of symmetric analysis and synthesis paths with four resolution levels each. Each resolution level in the analysis path contains two convolutional layers with $3\times3\times3$ kernels, each followed by ReLU activations and a $2\times2\times2$ max pooling with strides of two in each dimension. In the synthesis path, transposed convolutions are utilized to remap the lower resolution feature maps within the network to the higher resolution space of the input images. These
are again followed by two $3\times3\times3$ convolutions, each of which employs ReLU activations. Furthermore, 3D U-Net utilizes shortcut (or skip) connections from layers of equal resolution in the analysis path to provide higher-resolution features to the synthesis path \cite{cciccek20163d}. The final convolutional layer employs a voxel-wise \textit{softmax} activation function to compute a 3D probability map for each of the target organs as the output of our network.

In this example, we investigate a custom-build 3D FCN similar to 3D U-Net as illustrated in Fig. \ref{fig:3Dunet}. The network has the same input and output volume sizes and uses same size convolutions with zero padding throughout. For training, we use randomly cropped subvolumes extracted from several training CT volumes. Here, we chose a size of only $64\times 64\times 64$ for each subvolume that is small enough to allow the extraction of three subvolumes for mini-batch training on a single GPU. Batch sizes $>$1 can lead to better convergence during training by using batch normalization \cite{ioffe2015batch} when sampled from different patient volumes \cite{roth2017towards}. Our 3D FCN uses concatenation skip connections to the encoder part of the network as in the original 3D U-Net \cite{ronneberger2015unet,drozdzal2016importance}, resulting in $\sim$19M trainable parameters.

During inference (prediction), the network can be reshaped in order to process the test images more efficiently \cite{long2015fully}. Hence, the network is resized to an input size that covers the whole $xy$ dimension of a given CT volume and the full output is built up by applying an overlapping tiles approach in $z$-direction. Again, the network reshaping size is dependent on the amount of available GPU memory.
%%%%%%%%%%%%%%%%%%%%%%%%%%%%%%%%%%%%%%%%%%%%%%%%%%%%%%%%%%%%%%%
\subsection{Data augmentation} In training, we employ smooth B-spline deformations to both the image and label data, similar to \cite{cciccek20163d}. The deformation fields are randomly sampled from a uniform distribution with a maximum displacement of 4 and a grid spacing of 24 voxels. Furthermore, we applied random rotations between -$20^{\circ}$ and +$20^{\circ}$, and translations of -20 to +20 voxels in each dimension at each iteration in order to generate plausible deformations during training. This type of data augmentation can help to artificially increase the training data and encourages convergence to more robust solutions while reducing overfitting to the training data (see Fig. \ref{fig:training}). 
%%%%%%%%%%%%%%%%%%%%%%%%%%%%%%%%%%%%%%%%%%%%%%%%%%%%%%%%%%%%%%%
\subsection{Loss function}
The Dice similarity coefficient (DSC) is often used to measure the amount of agreement between two binary regions. Hence, it is widely used as a metric for evaluating performance of image segmentation algorithms. A differentiable version has been proposed by Milletari et al. \cite{milletari2016v} which we use for training our 3D U-Net model. In order to optimize the DSC score on the training data, we minimize the following loss function for each class $l$: 
\begin{equation}
\mathcal{L}_l=-\frac{2\sum_{i}^{N_v} p_ir_i}{\sum_{i}^{N_v} p_i+\sum_{i}^{N_v} r_i}.
\end{equation}
Here, $p_i$ represents the value of the \textit{softmax} probability map and $r_i$ the corresponding ground truth at voxel $i$ of $N_v$ in the current image volume. In order to predict multiple classes for segmentation, we calculate the total loss as
\begin{equation}
\mathcal{L}_\mathrm{total}=\frac{1}{L}\sum_{l}^{L} w_l L_l,
\end{equation}
where $L$ is the number of classes (number foreground classes, plus background) and $w_l$ is a weight factor that can influence the contribution of each label class $l$. In this example, we keep $w_l = 1$ for all labels. Alternative weighting schemes have been explored in \cite{shen2017dice,sudre2017generalised}. 
%%%%%%%%%%%%%%%%%%%%%%%%%%%%%%%%%%%%%%%%%%%%%%%%%%%%%%%%%%%%%%%%%%%%%%%%%%%%%%%%%%%%%%%%%%%%%%%%
\begin{figure}[htb]
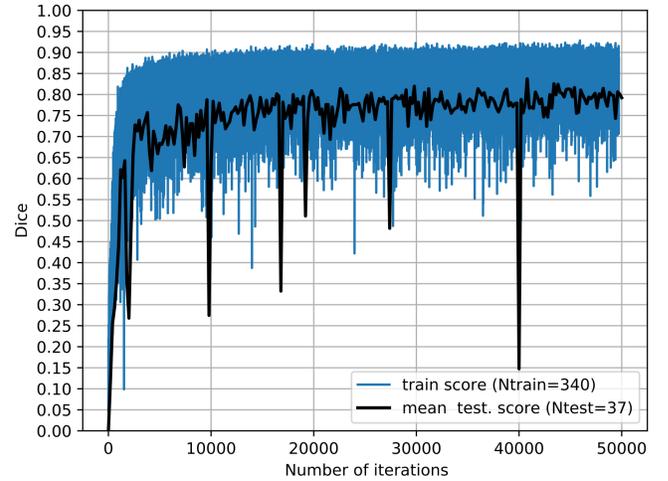

	\centering
	\adjincludegraphics[width=0.95\columnwidth]{acc_tf_down2_keras_unet_gtcrop64_minsum1_batchsz3_Dice_CONCAT--unet_pancreas_DSC.eps}
	\caption{Training/testing curves of the 3D FCN network for multi-organ segmentation using the Dice loss for optimization. \label{fig:training}}
\end{figure}
%%%%%%%%%%%%%%%%%%%%%%%%%%%%%%%%%%%%%%%%%%%%%%%%%%%%%%%%%%%%%%%%%%%%%%%%%%%%%%%%%%%%%%%%%%%%%%%%
\subsection{Implementation} All models are implemented in Keras\footnote{\url{https://keras.io}} with the TensorFlow\footnote{\url{https://www.tensorflow.org}} backend, which employs automatic differentiation in order compute the gradients for optimizing the model \cite{abadi2016tensorflow}. We use Adam optimization \cite{kingma2014adam} with an initial learning rate of $1\times10^{-2}$. We train our model for 50,000 iterations, which takes about one week on a NVIDIA Quadro P6000 GPUs with 24 GB. Inference, however, is achieved in less than 1 minute per case. 
%%%%%%%%%%%%%%%%%%%%%%%%%%%%%%%%%%%%%%%%%%%%%%%%%%%%%%%%%%%%%%%%%%%%%%%%%%%%%%%%%%%%%%%%%%%%%%%%
\section{EXPERIMENTS \& RESULTS}
%%%%%%%%%%%%%%%%%%%%%%%%%%%%%%%%%%%%%%%%%%%%%%%%%%%%%%%%%%%%%%%%%%%%%%%%%%%%%%%%%%%%%%%%%%%%%%%%
\begin{figure*}[tb]
	\centering
	\begin{tabular}{c c}
		\subfloat[Ground truth (axial)]{\rotatebox{180}{\adjincludegraphics[valign=c,width=0.35\textwidth]{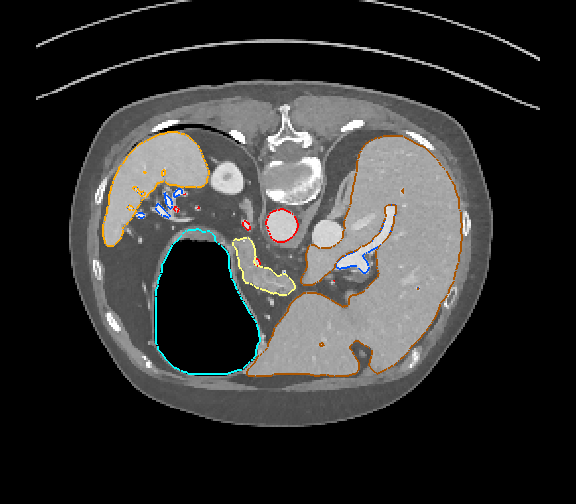}}} &
		%\hfill
		\subfloat[Ground truth (3D)]{\adjincludegraphics[valign=c,width=0.35\textwidth]{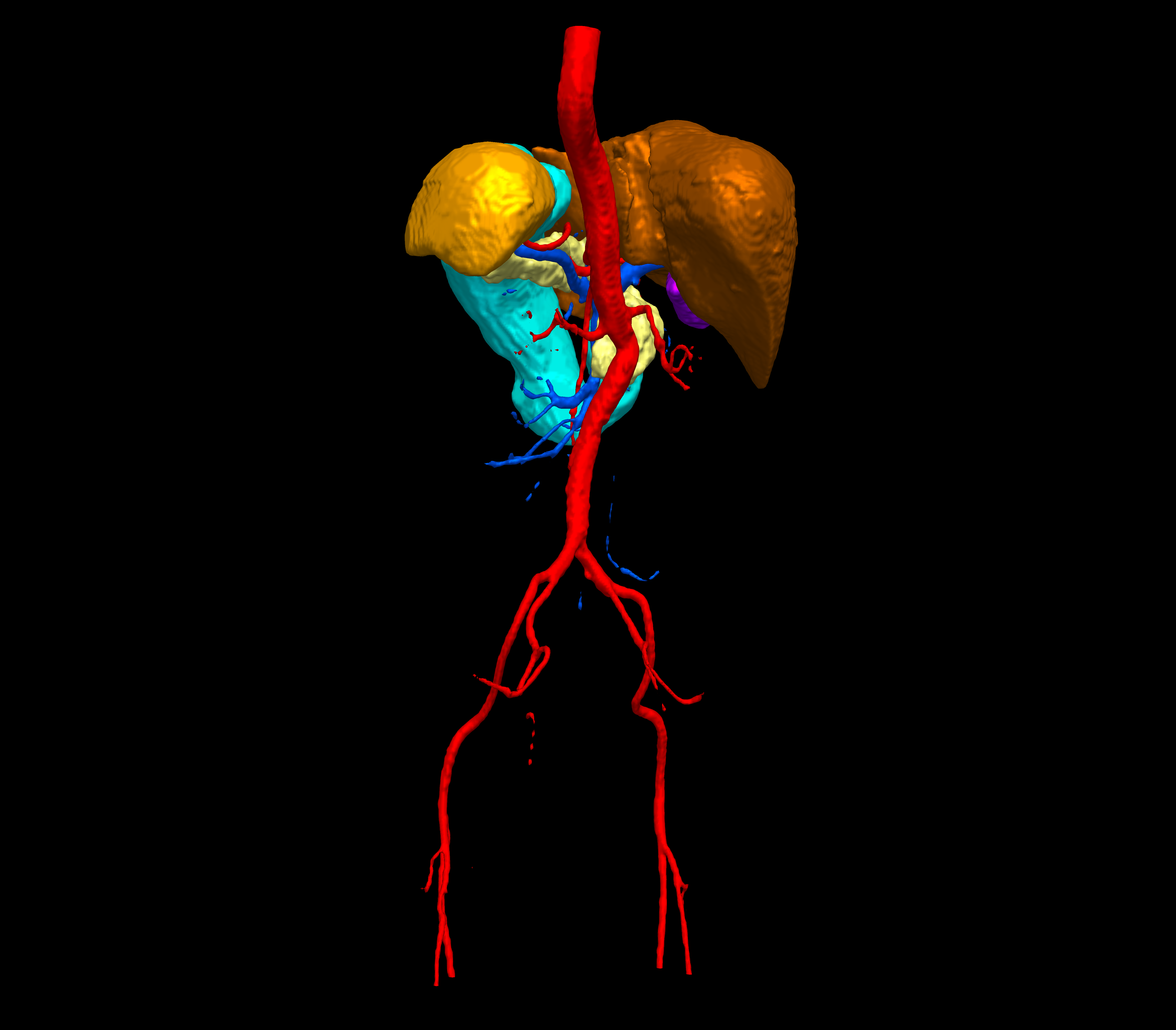}}  \\  
		%\hfill
		\subfloat[Segmentation (axial)]{\rotatebox{180}{\adjincludegraphics[valign=c,width=0.35\textwidth]{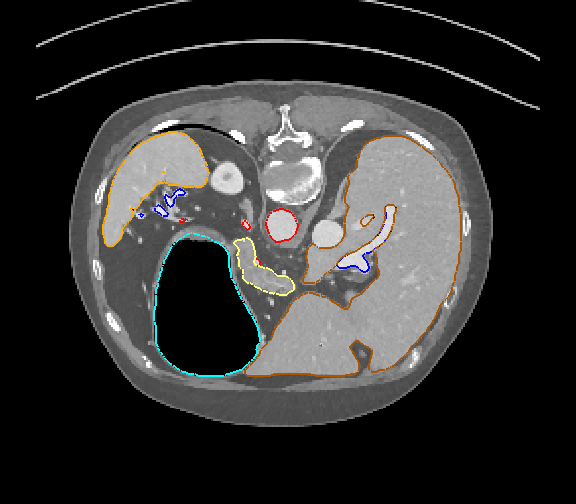}}}  &
		%\hfill
		\subfloat[Segmentation (3D)]{\adjincludegraphics[valign=c,width=0.35\textwidth]{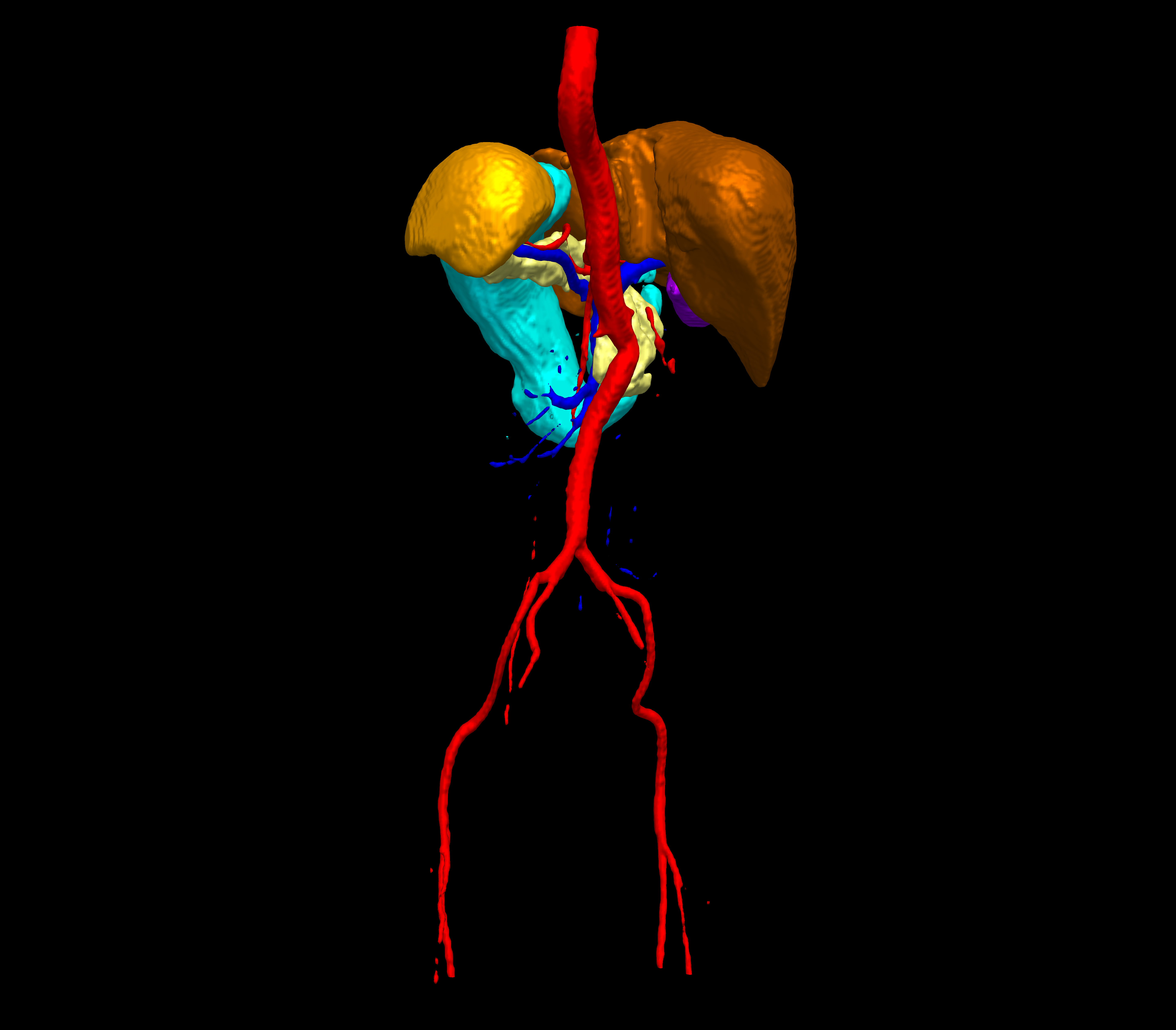}} 
	\end{tabular}
	\vspace{1em}
	\caption{Visualization of the resulting multi-organ segmentations in CT showing both axial (a, c) and 3D renderings (b, d) of the results and corresponding ground truth after up-sampling the prediction to half the original CT resolution.\label{fig:result_3D_view}}
\end{figure*}
%%%%%%%%%%%%%%%%%%%%%%%%%%%%%%%%%%%%%%%%%%%%%%%%%%%%%%%%%%%%%%%%%%%%%%%%%%%%%%%%%%%%%%%%%%%%%%%%
\subsection{Data}
Our dataset originates from a research study with gastric cancer patients. 377 contrast enhanced abdominal CT scans were acquired in the portal venous phase. Each CT volume consists of 460--1177 slices of $512\times512$ pixels. Voxel dimensions are [0.59-0.98, 0.59-0.98, 0.5-1.0] mm. In each image, the arteries, portal vein, liver, spleen, stomach, gallbladder, and pancreas were delineated by several trained researchers and confirmed by a clinician. Because of GPU memory constraints and computational efficiency, we downsample all images by a factor of 4, resulting axial images sizes of $128\times128\times$(number of slices)/4. 
%%%%%%%%%%%%%%%%%%%%%%%%%%%%%%%%%%%%%%%%%%%%%%%%%%%%%%%%%%%%%%%%%%%%%%%%%%%%%%%%%%%%%%%%%%%%%%%%
\begin{table}[t!]
	\small
	\centering
	\caption{Quantitative results of the 3D FCN network in \textbf{training} (n=340).}
	\label{tab:training}
	\begin{tabular}{lllll}
		\hline
		\textbf{Dice (\%)}   & \textbf{Avg.}   & \textbf{Std.}  & \textbf{Min.}   & \textbf{Max.}   \\
		\hline		
		\textbf{artery}      & 84.1\%          & 5.0\%          & 66.9\%          & 91.7\%          \\
		\textbf{vein}        & 77.5\%          & 8.9\%          & 29.2\%          & 89.2\%          \\
		\textbf{liver}       & 96.6\%          & 1.1\%          & 91.4\%          & 98.5\%          \\
		\textbf{spleen}      & 96.3\%          & 2.0\%          & 79.8\%          & 98.9\%          \\
		\textbf{stomach}     & 95.6\%          & 7.7\%          & 0.0\%           & 99.7\%          \\
		\textbf{gallbladder} & 90.1\%          & 10.9\%         & 0.0\%           & 97.8\%          \\
		\textbf{pancreas}    & 85.5\%          & 8.9\%          & 28.0\%          & 95.5\%          \\
		\hline		
		\textbf{Total Avg.}  & \textbf{89.4\%} & \textbf{6.4\%} & \textbf{42.2\%} & \textbf{95.9\%} \\
		\hline		
	\end{tabular}
\end{table}
%%%%%%%%%%%%%%%%%%%%%%%%%%%%%%%%%%%%%%%%%%%%%%%%%%%%%%%%%%%%%%%%%%%%%%%%%%%%%%%%%%%%%%%%%%%%%%%%
\begin{table}[t!]
	\small
	\centering
	\caption{Quantitative results of the 3D FCN network in \textbf{testing} (n=37).}
	\label{tab:testing}
	\begin{tabular}{lllll}
		\hline
		\textbf{Dice (\%)}   & \textbf{Avg.}   & \textbf{Std.}  & \textbf{Min.}   & \textbf{Max.}   \\
		\hline		
		\textbf{artery}      & 83.5\%          & 4.1\%          & 73.7\%          & 91.1\%          \\
		\textbf{vein}        & 80.5\%          & 6.8\%          & 49.0\%          & 89.4\%          \\
		\textbf{liver}       & 97.1\%          & 1.0\%          & 93.5\%          & 98.3\%          \\
		\textbf{spleen}      & 97.7\%          & 0.8\%          & 95.2\%          & 98.9\%          \\
		\textbf{stomach}     & 96.1\%          & 7.9\%          & 49.4\%          & 98.9\%          \\
		\textbf{gallbladder} & 85.1\%          & 15.7\%         & 28.6\%          & 97.4\%          \\
		\textbf{pancreas}    & 84.9\%          & 9.1\%          & 52.5\%          & 95.1\%          \\
		\hline		
		\textbf{Total Avg.}  & \textbf{89.3\%} & \textbf{6.5\%} & \textbf{63.1\%} & \textbf{95.6\%} \\
		\hline		
	\end{tabular}
\end{table}
%%%%%%%%%%%%%%%%%%%%%%%%%%%%%%%%%%%%%%%%%%%%%%%%%%%%%%%%%%%%%%%%%%%%%%%%%%%%%%%%%%%%%%%%%%%%%%%%
\subsection{Evaluation}
We evaluate this model using a random split of 340 training and 37 testing patients, and achieve an average Dice score performance of 89.4 $\pm$ 6.4 (range [42.2, 95.9])\% in training (see Table \ref{tab:training}), and 89.3 $\pm$ 6.5 (range [63.1, 95.6])\% in testing (see Table \ref{tab:testing}). This result indicates that the used dataset size and use of data augmentation is sufficient to avoid overfitting to the training data. Furthermore, this result is comparable or better than other state-of-the-art deep learning architectures for single and multi-organ segmentation in CT \cite{roth2017towards,gibson2017towards,zhou2017deep,roth2017hierarchical}. However, direct comparison to other methods is difficult due to the different datasets and validation schemes employed. An example prediction result of the model is shown in Fig. \ref{fig:result_3D_view}.
%%%%%%%%%%%%%%%%%%%%%%%%%%%%%%%%%%%%%%%%%%%%%%%%%%%%%%%%%%%%%%%%%%%%%%%%%%%%%%%%%%%%%%%%%%%%%%%%
%%%%%%%%%%%%%%%%%%%%%%%%%%%%%%%%%%%%%%%%%%%%%%%%%%%%%%%%%%%%%%%%%%%%%%%%%%%%%%%%%%%%%%%%%%%%%%%%
\section{CONCLUSIONS}
This example model achieves state-of-the-art performances in automated multi-organ segmentation of abdominal CT with $\sim$90\% average Dice score in testing across all targeted organs. We showed that deep 3D FCNs can be efficiently trained on modern GPUs. In the future, the availability of larger amounts of GPU memory will allow the processing of whole CT volumes at higher resolution. Increasing dataset sizes will likely further improve the performance of automated multi-organ segmentation in medical imaging. The current model does not contain any constraints on the shape of the segmented anatomy and can result in disconnected or isolated regions. In the future, some anatomical constraints could be included in order to guarantee topologically correct segmentation results \cite{oktay2017anatomically}.
% The utilization of summation layers rather than concatenation layers seems to be beneficial for certain applications such as pancreas segmentation.
%%%%%%%%%%%%%%%%%%%%%%%%%%%%%%%%%%%%%%%%%%%%%%%%%%%%%%%%%%%%%%%%%%%%%%%%%%%%%%%%%%%%%%%%%%%%%%%%
%%%%%%%%%%%%%%%%%%%%%%%%%%%%%%%%%%%%%%%%%%%%%%%%%%%%%%%%%%%%%%%%%%%%%%%%%%%%%%%%%%%%%%%%%%%%%%%%
\section*{Acknowledgments}
\small This research was supported by MEXT KAKENHI (26108006, 26560255, 25242047, 17H00867, 15H01116) and the JPSP International Bilateral Collaboration Grant. 
%%%%%%%%%%%%%%%%%%%%%%%%%%%%%%%%%%%%%%%%%%%%%%%%%%%%%%%%%%%%%%%%%%%%%%%%%%%%%%%%%%%%%%%%%%%%%%%%
%%%%%%%%%%%%%%%%%%%%%%%%%%%%%%%%%%%%%%%%%%%%%%%%%%%%%%%%%%%%%%%%%%%%%%%%%%%%%%%%%%%%%%%%%%%%%%%%
%\clearpage
%\newpage
\small
\bibliographystyle{ieeetr}  % (1) prints author names abbreviated (like {abbrv}) in the references section and (2) sorts references numerically in citation order (like {unsrt})
\bibliography{rothhr_references}
\vspace{1em}
%%%%%%%%%%%%%%%%%%%%%%%%%%%%%%%%%%%%%%%%%%%%%%%%%%%%%%%%%%%%%%%%%%%%%%%%%%%%%%%%%%%%%%%%%%%%%%%%
%% Author biographies
%%%%%%%%%%%%%%%%%%%%%%%%%%%%%%%%%%%%%%%%%%%%%%%%%%%%%%%%%%%%%%%%%%%%%%%%%%%%%%%%%%%%%%%%%%%%%%%%%%%%%%%%%%%%%%%%%%%%%%%%%%%
\noindent\parbox{\columnwidth}{
\setlength\intextsep{0pt}
\begin{wrapfigure}{l}{25mm} 
 	\includegraphics[width=0.15\textwidth,keepaspectratio]{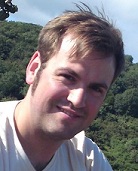}
\end{wrapfigure}\par\noindent\textbf{Holger R. Roth} is Assistant Professor (Research) at Nagoya University. His focus is the application of deep learning methods for medical image analysis and improved computer-aided diagnosis. He received his PhD at the Centre for Medical Image Computing, University College London, UK, and was a Research Fellow at the Imaging Biomarkers and CAD Lab at the National Institutes of Health, USA, from 2013 to 2016.\par
\vspace{1em}
}
\vfill\null
%\columnbreak
%%%%%%%%%%%%%%%%%%%%%%%%%%%%%%%%%%%%%%%%%%%%%%%%%%%%%%%%%%%%%
\noindent\parbox{\columnwidth}{
\setlength\intextsep{0pt}
\begin{wrapfigure}{l}{25mm} 
	\includegraphics[width=0.15\textwidth,keepaspectratio]{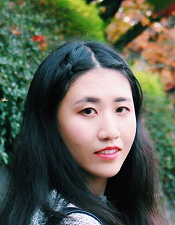}
\end{wrapfigure}\par\noindent\textbf{Chen Shen} is a Master's student at Nagoya University, Japan. Her major is in Intelligent Systems. Her research interests lie in organ segmentation using deep learning methods. She graduated from East China University of Science and Technology, China, in 2016.\par
\vspace{6em}
}
%%%%%%%%%%%%%%%%%%%%%%%%%%%%%%%%%%%%%%%%%%%%%%%%%%%%%%%%%%%%%
\noindent\parbox{\columnwidth}{
\setlength\intextsep{0pt}
\begin{wrapfigure}{l}{25mm} 
	\includegraphics[width=0.15\textwidth,keepaspectratio]{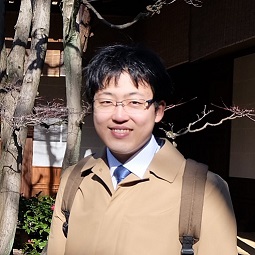}
\end{wrapfigure}\par\noindent\textbf{Hirohisa Oda} received a MEng degree from Nagoya University, Japan, in 2014. After working in industry, he started a PhD program of Nagoya University from 2015. His specializations are image processing and machine learning. His research interests are computer-aided diagnosis and microfocus X-ray CT for cardiac image processing.\par
\vspace{1em}
}
%%%%%%%%%%%%%%%%%%%%%%%%%%%%%%%%%%%%%%%%%%%%%%%%%%%%%%%%%%%%%
\noindent\parbox{\columnwidth}{
\setlength\intextsep{0pt}
\begin{wrapfigure}{l}{25mm} 
	\includegraphics[width=0.15\textwidth,keepaspectratio]{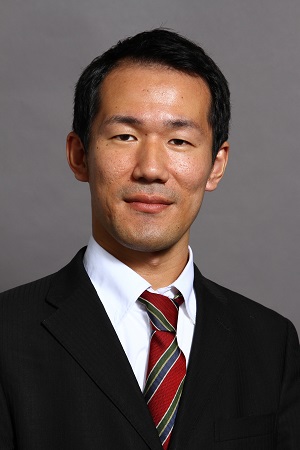}
\end{wrapfigure}\par\noindent\textbf{Masahiro Oda} graduated from the Department of Information Engineering, School of Engineering, Nagoya University in 2004. In 2009, he completed a doctor’s course at the Department of Media Science, Graduate School of Information Science, Nagoya University were he is now assistant professor of the Graduate School of Informatics, Nagoya University. He received a RSNA Certificate of Merit in 2009, a RSNA Magna Cum Laude in 2014, and some other awards. His research fields are computer-aided diagnosis and surgery using medical image processing. He is a member of The Institute of Electronics, Information and Communication Engineering, The Japan Society of Medical Imaging Technology, Japan Society of Computer Aided Surgery, and Information Processing Society of Japan.\par
\vspace{1em}
}
%%%%%%%%%%%%%%%%%%%%%%%%%%%%%%%%%%%%%%%%%%%%%%%%%%%%%%%%%%%%%
\noindent\parbox{\columnwidth}{
	\setlength\intextsep{0pt}
	\begin{wrapfigure}{l}{25mm} 
		% none
	\end{wrapfigure}\par\noindent\textbf{Yuichiro Hayashi}\par
	\vspace{1em}
}
%%%%%%%%%%%%%%%%%%%%%%%%%%%%%%%%%%%%%%%%%%%%%%%%%%%%%%%%%%%%%
\noindent\parbox{\columnwidth}{
\setlength\intextsep{0pt}
\begin{wrapfigure}{l}{25mm} 
	\includegraphics[width=0.15\textwidth,keepaspectratio]{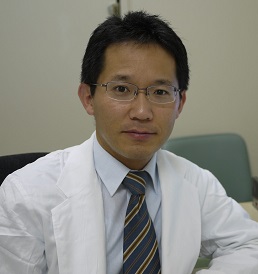}
\end{wrapfigure}\par\noindent\textbf{Kazunari Misawa, MD, PhD}, is Clinical Director in the Department of Gastroenterological Surgery, Aichi Cancer Center Hospital. He specializes in surgical oncology for gastric cancer, laparoscopic surgery, and reduced port surgery. His research covers minimally invasive surgery, postoperative QOL, computer aided surgery. He received his MD from Nagoya University, Nagoya, Japan in 1995,  and a PhD in Medical science in 2014.\par
\vspace{1em}
}
%%%%%%%%%%%%%%%%%%%%%%%%%%%%%%%%%%%%%%%%%%%%%%%%%%%%%%%%%%%%%
\noindent\parbox{\columnwidth}{
	\setlength\intextsep{0pt}
	\begin{wrapfigure}{l}{25mm} 
		\includegraphics[width=0.15\textwidth,keepaspectratio]{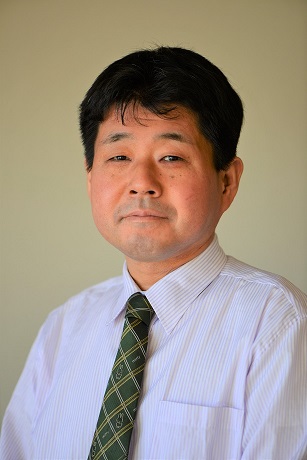}
	\end{wrapfigure}\par\noindent\textbf{Kensaku Mori} is a professor of Information and Communications and Graduate School of Information Science, Nagoya University and the director of Information Technology Center of Nagoya University. He received MEng (Information Engineering) and PhD (Information Engineering) from Nagoya University in 1994 and 1996. He started his academic carrier in 1997 as an assistant professor of Nagoya University. Then he was promoted as associate professor in 2001 in Nagoya University. In 2009, he has obtained full professorship in Nagoya University. In 2016, he was appointed as the director of Information Technology Center of Nagoya University. In 2017, he was appointed as a professor of Department of Intelligent Systems, Graduate School of Informatics.\par
	\vspace{1em}
}
%%%%%%%%%%%%%%%%%%%%%%%%%%%%%%%%%%%%%%%%%%%%%%%%%%%%%%%%%%%%%%%%%%%%%%%%%%%%%%%%%%%%%%%%%%%%%%%%
\end{document}